# An Evolutionary Squeaky Wheel Optimisation Approach to Personnel Scheduling

Uwe Aickelin, Edmund K. Burke and Jingpeng Li*

School of Computer Science
The University of Nottingham
Nottingham, NG8 1BB
United Kingdom
{uxa, ekb, jpl}@cs.nott.ac.uk
*Authors are in alphabetical order. Please send all correspondence to Jingpeng Li

Abstract. The quest for robust heuristics that are able to solve more than one problem is ongoing. In this paper, we present, discuss and analyse a technique called Evolutionary Squeaky Wheel Optimisation and apply it to two different personnel scheduling problems. Evolutionary Squeaky Wheel Optimisation improves the original Squeaky Wheel Optimisation's effectiveness and execution speed by incorporating two additional steps (Selection and Mutation) for added evolution. In the Evolutionary Squeaky Wheel Optimisation, a cycle of Analysis-Selection-Mutation-Prioritization-Construction continues until stopping conditions are reached. The aim of the Analysis step is to identify below average solution components by calculating a fitness value for all components. The Selection step then chooses amongst these underperformers and discards some probabilistically based on fitness. The Mutation step further discards a few components at random. Solutions can become incomplete and thus repairs may be required. The repair is carried out by using the Prioritization step to first produce priorities that determine an order by which the following Construction step then schedules the remaining components. Therefore, improvements in the Evolutionary Squeaky Wheel Optimisation is achieved by selective solution disruption mixed with iterative improvement and constructive repair. Strong experimental results are reported on two different domains of personnel scheduling: bus and rail driver scheduling and hospital nurse scheduling.

## 1 Introduction

One of the more recent goals of heuristic research is to find algorithms that are robust enough to cope with a range of problem instances or ideally even different problem scenarios, without requiring re-development, intensive retraining or parameter tweaking. We address this issue by trying to design a heuristic that works equally well for two related, but very different personnel scheduling problems. Personnel scheduling problems have been addressed by managers, operational researchers and computer scientists over the past forty years. Over these years, there has been a wealth of literature on automated personnel scheduling including several survey papers that general-ise the problem classification and the associated approaches (Tien and Kamiyama, 1982; Ernst et al, 2004).

In brief, personnel scheduling is the problem of assigning staff members to shifts or duties over a scheduling period (typically a week or a month) so that certain constraints (legal, organizational and personal) are satisfied. Broadly speaking, there are two classes of personnel scheduling problems: cyclical and non-cyclical schedules. In cyclical scheduling, the same set of patterns is always used, but rotated amongst employees. Cyclical schedules have fallen out of fashion in recent years, because they are too restrictive and do not allow personal preferences to be included. Hence, we concentrate in this paper on two non-cyclical scheduling problems.

The non-cyclical scheduling process normally consists of three stages: the first stage involves determining how many staff must be employed in order to meet the service demand; the second stage involves allocating individual staff members to overall shifts patterns and the final stage assigns actual duties to individuals for each shift. Throughout the process, all industrial regulations associated with the relevant workplace agreements must be complied with. An Integer Linear Programme (ILP) was first proposed by Dantzig (1954) to formulate general personnel scheduling problems as follows.

Objective Function:
$$\text{Minimize} \sum_{i \in I} c_i x_i \tag{1}$$

Subject to:
$$\sum_{i \in I} a_{ti} x_i \geq b_t, \quad t \in T \tag{2}$$

$$x_i \in N, \quad i \in I \tag{3}$$

where I represents the set of all shifts, T represents the planning periods that the shift schedule covers, $c_i$ is the cost of assigning an employee to shift i, $b_t$ is the estimated demand for period t, $a_{ti}$ is 1 if period t is a work period for shift i and 0 otherwise, and $x_i$ is an integer variable defined as the number of employees assigned to shift j.

Equation (1) is the target function of minimizing total scheduling cost. Equation (2) is the coverage constraint which ensures the demand must be satisfied (in terms of number of employees needed), and Equation (3) ensures the integrality of variable $x_i$.

Personnel scheduling has important application in many problems such as crew scheduling in transportation systems (Bodin et al, 1983), nurse scheduling in health care systems (Burke et al, 2004), and tour scheduling (Bechtold et al, 1991) for various service systems. On all of these problems, there has been a wide variety of search methods and techniques such as mathematical programming (Gamache and Soumis, 1998), constraint programming (Guerinik and Van Caneghem, 1995; Cheng et al, 1997), fuzzy sets (Teodorovic 1998; Li and Kwan, 2003), expert systems (Anantaram 1993; Gray et al, 1993), hyper-heuristics (Burke et al, 2003) and meta-heuristics (see remainder of this section), of which meta-heuristics have attracted particular attention in recent years.

Genetic Algorithms form an important class of meta-heuristics, and have been extensively applied to personnel scheduling problems (Wren and Wren, 1995; Aickelin and Dowsland, 2000; Cai and Li, 2000; Kwan et al, 2001; Aickelin and White, 2004;). A number of attempts have also been made using other meta-heuristics, such as tabu search (Cavique et al, 1999; Burke et al, 1999; Shen and Kwan, 2001), simulated annealing (Brusco and Jacobs, 1993; Abramson 1996), memetic algorithms (Burke et al, 2001), Bayesian optimization (Li and Aickelin, 2004), ant colony optimization (Aickelin et al, 2007; Gutjahr and Rauner, 2007) and variable neighbourhood search (Burke et al, 2007). The methods and techniques that have been used over the years to tackle personnel scheduling problems have tended to draw on problem-specific information and particular heuristics. As a result, many heuristic methods are highly successful at solving particular problem instances, but cannot directly be applied to other problem instances or related problems (Randhawa and Sitompul, 1993; Kwan et al, 2001). In this paper, we are trying to develop more general personnel scheduling systems, i.e. our goal is to find a method that is not designed with one particular problem in mind, but is instead applicable to a range of problems and domains.

The work that is presented here is based on the observation that in most real world problems the solutions consist of components which are intricately woven together in a non-linear, non-additive fashion. Each solution component, e.g. a shift pattern assigned to a particular employee, may be a strong candidate in its own right, but it also has to fit well with other components. To deal with these components, Joslin and Clements (1999) proposed a technique called Squeaky Wheel Optimisation, and claimed it could be used to solve various combinatorial optimisation problems. In this paper, we analyse the limitations of the original technique and revise it by incorporating some evolutionary features into the searching process. We term the revised version the Evolutionary Squeaky Wheel Optimisation method. The basic idea behind the Evolutionary Squeaky Wheel Optimisation was initially and briefly presented in Aickelin et al (2006) for the driver scheduling problem. Here we present it in more detail and show that it is indeed as robust as we claimed by solving different problems. Its general idea is to break a solution down into its components and assign a score to each. The scores are used in two ways: first as fitness values which determine the probability for the components to survive in the current solution, and then as an index to obtain an order in which a greedy algorithm reschedules deleted components.

The next section describes the characteristics of the Evolutionary Squeaky Wheel Optimisation that make it an efficient technique for search. The following two sections discuss implementations of the technique on two different domains of personnel scheduling: the bus and rail driver scheduling in transportation systems and the hospital nurse scheduling in healthcare systems, respectively. The final section gives some concluding remarks and discusses how far we have come on the road towards robust heuristics.

## 2 A General Description of the Evolutionary Squeaky Wheel Optimisation Method

A detailed description of the Evolutionary Squeaky Wheel Algorithm can be found in Aickelin et al (2006). What follows here is a shorter summary to aid in the understanding of the remainder of this paper. Squeaky Wheel Optimization belongs to the class of non-systematic search techniques. In Squeaky Wheel Optimisation, a priority ordering of problem components is given to a greedy algorithm that constructs a solution. That solution is then analyzed to find components that are not handled as well as they could be, relative to some lower bound. The priority to be re-scheduled of these components is then increased. Hence the name of the heuristic is derived from the common saying 'Find the most squeaky wheel and then oil it'.

All components, sorted in the new priority ordering are given to a greedy constructor, with the likely result that 'squeaky' components will be handled better in the next solution. This construct-analyze-prioritize cycle continues until a stopping condition is reached. Joslin and Clements (1999) applied this technique on production line scheduling problems and graph colouring problems with some satisfactory results. Burke and Newall (2004) developed an adaptive heuristic framework for examination timetabling problems that was based on Squeaky Wheel Optimisation. A hybridisation of this method with an exam timetabling methodology based upon the Great Deluge algorithm was shown to be effective on benchmark problems.

In essence, Squeaky Wheel Optimisation finds good quality solutions quickly by searching in two spaces simultaneously: the traditional solution space and the new priority space. Hence, it avoids many problems that other local search methods often encounter. These features allow Squeaky Wheel Optimisation to effectively make large coherent moves to escape from unpromising regions in the search space. The construct-analyze-prioritize loop has the consequence that problem components that are hard to handle tend to rise in the priority queue, and components that are easy to handle tend to sink.

Squeaky Wheel Optimisation suffers from two limitations: poor scalability due to the random starting point of the greedy constructor and slow convergence due to the inability to make very small moves in the solution space. The second weakness is caused by the heuristic operating on dual search spaces (one of its "strengths"), i.e. a small change in the sequence of components generated by the Prioritization step may correspond to a large change in the corresponding solution generated by the Construction step.

To address these weaknesses we have developed the Evolutionary Squeaky Wheel Algorithm (Aickelin et al 2006): We believe that if the construction process started from partial solutions that contain information of past solutions, the optimisation process would speed up significantly. Moreover, if it was possible to restrict changes of components to the trouble-makers only, e.g. by delaying part of the sequence without going through the full Analysis and Prioritization cycle, then the changes in the corresponding solutions would be relatively small (Aickelin et al 2006).

Evolutionary Squeaky Wheel Optimisation incorporates two additional steps into the heuristic loop: Selection and Mutation. Of key importance is that the admittance of a new component is analyzed by a dynamic evaluation function, which takes ac-count of how well the prospective component fits with others already in the solution. We give the pseudo-code of the algorithm in Figure 1.

```
The Evolutional Squeaky Wheel Optimization Algorithm
{
  /*
  E={e₁,…,eₘ}: Set of all available components;
  S={s₁,…,sₙ}: Subset of E which includes the components used in
    a Current solution;
  F(sᵢ): Fitness of sᵢ;
  Q[]: Queue to store the deleted components;
  P_M: Mutation rate;
  Analyze(sᵢ,S): Function of analyzing sᵢ in S;
  Assign(eᵢ',E,Q[i]): Function of assigning eᵢ' in E to fulfill
    the task of the deleted component stored at Q[i];
  */
  Repeat {
    Analysis:
      For (i=0;i<|S|;i++) F(sᵢ)=Analyze(sᵢ,S);
    Selection:
      For (i=0;i<|S|;i++)
        If (random>F(sᵢ) {Add sᵢ to Q; Delete sᵢ from S;}
    Mutation:
      For (i=0;i<|S|;i++)
        If (random<p_M) {Add sᵢ to Q; Delete sᵢ from S;}
    Prioritization:
      Sort Q in the ascending order of the Fᵢ values of sᵢ;
    Construction:
      For (i=0;i<|Q|;i++) {Assign(eᵢ',E,Q[i]); Add eᵢ' to S;}
      Empty Q;
  }
  Until Stopping condition is satisfied;
  Return the best solution;
}
```

Figure 1: General structure of the Evolutionary Squeaky Wheel algorithm

## 3 Evolutionary Squeaky Wheel Optimisation for Driver Scheduling

### 3.1 Problem description

The first of the two problem classes we apply the Evolutionary Squeaky Wheel Optimisation to is bus and rail driver scheduling, which has attracted much interest since the 1960's (see the survey articles in Wren and Rousseau, 1995; Kwan 2004). Driver scheduling represents the process of partitioning blocks of work, each serviced by one vehicle, into a set of legal driver shifts (Wren and Wren 1995). To clarify the problem, we start from some terms introduced in (Li and Kwan, 2003). Relief Opportunities are time/location pairs at which drivers can change vehicles, and the individual period between any two successive relief opportunities on a particular vehicle is known as a piece of work. The work of a driver in a day is known as a shift, which is composed of several spells of work each of which contains a number of consecutive pieces of work on the same vehicle. A schedule is defined as a solution that contains a set of shifts that cover all the required work, with the smallest number of drivers and the least costs.

The driver scheduling problem can be formulated as the following set covering integer programming problem which ensures that all of the vehicle work is covered. Computational difficulty exists with this approach due to the enormous size of the search spaces generated.

Decision variables:
$X_i$ =1 if shift i is used in the solution, 0 otherwise.

Parameters:
m = Number of potential shifts;
n = Number of work pieces to be covered;
$c_i$ = Cost of shift i (hours paid);
$a_{ij}$ = 1 if shift i covers work piece j, 0 otherwise.

Objective function:

$$\text{Minimize} \sum_{i=1}^{n} c_i x_i \qquad (4)$$

Subject to:

$$\sum_{i=1}^{m} a_{ij} x_i \geq 1, j \in \{1, \ldots, n\} \qquad (5)$$

$$x_i \in \{0,1\}, i \in \{1, \ldots, m\} \qquad (6)$$

Objective (4) minimizes the total cost. Constraint (5) ensures that each piece of work is covered by at least one driver, and constraint (6) requires that whole shifts be considered.

3.2 Implementation

This section details how to apply the Evolutionary Squeaky Wheel Optimisation to driver scheduling. The steps of Analysis, Selection, Mutation, Prioritization and Construction are executed repeatedly to improve a given initial solution. During each iteration, an unfit portion of the working schedule is removed and the resulting partial schedule is repaired by the constructing heuristic. The best schedule obtained is returned as the final solution.

3.2.1 Analysis
This step is to evaluate the current arrangements of individual shifts in a schedule by their fitness values, which can be formulated by employing the following normalized function

$$F(s_i) = f_1(s_i) \times f_2(s_i), \forall i \in \{1,...,|S|\} \quad (7)$$

where $s_j$ denotes the shift contained in the current schedule S with an index i, $f_1(s_i)$ is the structural coefficient of shift $s_i$, and $f_2(s_i)$ is the over-cover penalty reflecting the coverage status of shift $s_i$.

1) Structural coefficient

Five fuzzified criteria $u_j$ (j = 1,..., 5), characterized by associated membership functions, have been set up for the evaluation of the shift structure (Li and Kwan 2003). They are the actual work-time $u_1$, the ratio $u_2$ of actual work-time to paid hours, the number of pieces of work $u_3$, the number of spells $u_4$ contained in a shift, and the fractional cover $u_5$ given by a linear programming relaxation. Since each criterion evaluates the shift structure from different aspect, an overall evaluation (i.e. the calculation of the structural coefficient $f_1(s_i)$ for shift $s_i$) could be made by aggregating these five criteria as

$$f_1(s_i) = \sum_{k=1}^{5} w_k \mu_{\tilde{A}_i}, \forall i \in \{1,...,|S|\} \quad (8)$$

where $\tilde{A}_k$ is the fuzzy subset on the k-th criterion and $w_k$ is the weight of criterion $u_k$, s.t.

$$\sum_{k=1}^{5} w_k = 1, w_k \geq 0 \quad (9)$$

The design of the membership functions corresponding to these five criteria are briefly described as follows. As the efficiency of shift $s_i$ generally increases with the actual work-time, the ratio of actual work-time to paid hours and the number of pieces of work, respectively, the membership functions $\mu_{\tilde{A}_i}$ (k = 1, 2, 3) for these three criteria take the same form as

$$\mu_{\tilde{A}_i} = \begin{cases} 2\left(\dfrac{\lambda_i - b_i}{a_i - b_i}\right)^2, & a_i \leq \lambda_i < \dfrac{a_i + b_i}{2} \\ 1 - 2\left(\dfrac{\lambda_i - a_i}{a_i - b_i}\right)^2, & \dfrac{a_i + b_i}{2} \leq \lambda_i \leq a_i \end{cases} \quad (10)$$

where $\mu_{\tilde{A}_1}$ is the actual work-time of $s_i$, $\mu_{\tilde{A}_2}$ is the ratio of actual work-time to paid hours for $s_i$, $\mu_{\tilde{A}_3}$ is the number of pieces of work contained in $s_i$, $a_i$ (k = 1, 2, 3) are the maxima and $b_i$ (k = 1, 2, 3) are the minima.

With regard to criterion $u_4$, in most practical problems, the number of spells in a shift is limited to be at most four. From the viewpoint of transport operators, a 2-spell shift is generally the most effective, and a 3-spell shift is more desirable than the rest. Hence, we define the membership function $\mu_{\tilde{A}_4}$ as

$$\mu_{\tilde{A}_4} = \begin{cases} 0, & \text{if } \lambda_4 = 1 \text{ or } \lambda_4 = 4 \\ 0.5, & \text{if } \lambda_4 = 3 \\ 1, & \text{if } \lambda_4 = 2 \end{cases} \quad (11)$$

where $\mu_{\tilde{A}_4}$ is the number of spells in $s_i$.

With regard to the last criterion $u_5$, extensive studies have shown that the fractional cover by linear programming relaxation provides some useful information about the significance of some of the shifts identified in the relaxed solution. In general, the higher the fractional value of the variable for a shift, the higher chance that it would be present in the integer solution (Kwan et al, 2001). We use the following Gaussian distribution function $\mu_{\tilde{A}_5}$ to define criterion $u_5$ (see (Li and Kwan, 2003) for more details).

$$\mu_{\tilde{A}_5} = \begin{cases} e^{\frac{\ln 0.01}{(a_5-b_5)^2}(\lambda_5-a_5)^2}, & \text{if } s_i \text{ is in the fractional cover} \\ 0, & \text{otherwise} \end{cases} \quad (12)$$

where $\mu_{\tilde{A}_5}$ is the fractional value of $s_i$ in the relaxed LP solution, $a_5$ and $b_5$ are the maximum and the minimum respectively.

2) Over-cover penalty

Over-cover penalty $f_2(s_i)$, i.e. the ratio of overlapped work time to total work time in $s_j$, is the second part of the evaluation function (7), which can be formulated as

$$f_2(s_i) = \sum_{j=1}^{|s_i|}(\alpha_{ij} \times \beta_{ij}) \bigg/ \sum_{j=1}^{|s_i|}\beta_{ij}, \quad \forall i \in \{1,...,|S|\} \quad (13)$$

where $|s_i|$ is the number of pieces of work contained in $s_i$, $\alpha_{ij}$ is 0 if work piece j in $s_i$ has been covered by any other shifts $s_i$ in S and 1 otherwise, and $\beta_{ij}$ is the work-time for work pieces j in $s_i$.

3.2.2 Selection

This step is to decide the suitability for a shift to remain in a current schedule by comparing its fitness value $F(s_i)$ to $(p_s - p)$, where $p_s \in [0, 1]$ is a random number generated for each iteration and $p \in [0, 1]$ is a constant. The introduction of p is to prevent all shifts in the schedule from being removed while $p_s$ is close to 1, thus improving the Selection efficiency. If $F(s_i) \geq (p_s - p)$, then $s_i$ will stay at its present allocation. Otherwise, $s_i$ will be removed and the pieces of work it covers are released if they are not covered by the other shifts in the schedule. As the result of Selection, fitter shifts have more chances to survive in the current schedule.

3.2.3 Mutation

This step further mutates the remaining shifts in the partial schedule resulting from Selection, i.e. randomly discarding them at a constant rate of $p_M$. The pieces of work that the mutated shifts cover are released if they are not covered by the remaining shifts. Note that the mutation rate $p_M$ should be small to ensure convergence.

### 3.2.4 Prioritization

The discarded shifts could be regarded as troublesome ones due to their unfitness evaluated by Selection or unluckiness caused by Mutation, and a good rescheduling strategy for them is vital. Using the results of Analysis, the Prioritization step first produces a sequence of the troublesome shifts by sorting them in ascending order according to their fitness values, with ill-fitted shifts being earlier in the sequence.

The Prioritization step then uses the following technique to determine the order in which a new solution is constructed. Since each shift consists of several pieces of work, the sequence of shifts can be mapped to a longer sequence of pieces of work, with pieces that have already been covered by earlier shifts not appearing again. Thus, the Prioritization produces a new sequence consisting of all the uncovered pieces of work, in the order that they would be covered by the next step (Construction).

### 3.2.5 Construction

This step assigns new shifts sequentially to cover all uncovered pieces of work. The greedy constructor assumes that the desirability of adding a shift $s_i$ into the partial schedule increases with its function value $F(s_i)$. Considering all potential shifts with respect to the uncovered pieces of work, we need to build a coverage list for each piece containing the shifts that are able to cover it. The constructing process can be briefly described as follows: locate the first piece of work appearing in the sequence given by Prioritization, obtain its corresponding coverage list, randomly select a shift with the k-largest value $F(s_i)$, and update the coverage list and the sequence of the pieces of work. For a feasible solution obtained in such a way, there certainly exists over-cover, which can be easily removed by manual editing before the schedule is implemented.

The evaluation function used in Construction would take the same form as in Analysis. However, its evaluation is made on a different group of individuals for a different purpose: in Construction we evaluate the full set of potential shifts to add new shifts in completing an ideal schedule, while in Selection we evaluate a much smaller set which only consists of shifts in a current schedule to make deletion.

### 3.3 Experimental results

Among various heuristic and meta-heuristic approaches developed in recent years for driver scheduling, the Self-Adjusting Approach (SAA) performs generally best on a set of standard test instances (Li and Kwan, 2005). It uses the following weighted-sum objective function to combine the two objectives of minimizing the number of shifts and the total cost:

$$\text{Minimize} \sum_{k=1}^{K} (c_k + 2000) \tag{14}$$

where K is the number of shifts in the schedule, $c_k$ is the cost of the k-th shift measured in minute and an additional 2000-minute is used to give priority to the first objective of minimizing the number of shifts.

For a benchmark comparison, we use the same objective function. Our Evolutionary Squeaky Wheel Optimisation is coded in C++ and all experiments are imple-mented on the same Pentium IV 2.1 GHz machine. For all the thirteen test instances, we set the stopping criterion to be 1000 iterations without further improvement, and set the weight vector $w_i$ in (9) to be (0.20, 0.10, 0.10, 0.20, 0.40). In addition, we set $p_s$ in Section 3.2.2 to be 0.3, $p_M$ in Section 3.2.3 to be 0.05, and k in Section 3.2.5 to be 2. For each instance, we run the programme ten times from different random seeds.

| Data | Size | Shifts | IP Shift | IP Cost (hours) | IP CPU (sec) | SAA Shift Best | SAA Shift Mean | SAA Cost Best | SAA Cost Mean | SAA CPU (sec) |
|---|---|---|---|---|---|---|---|---|---|---|
| Bus1 | 127 | 3560 | 34 | 288 | 22 | 35 | 36 | 294 | 294 | 28 |
| Bus2 | 154 | 11817 | 34 | 289 | 84 | 35 | 35 | 294 | 295 | 26 |
| Bus3 | 613 | 22568 | 75 | 851 | 452 | 74 | 75 | 830 | 830 | 216 |
| Train1 | 340 | 29380 | 62 | 509 | 955 | 62 | 63 | 507 | 508 | 131 |
| Train2 | 437 | 25099 | 116 | 1004 | 69 | 117 | 117 | 998 | 1003 | 167 |
| Train3 | 456 | 16636 | 50 | 403 | 34 | 51 | 51 | 406 | 407 | 11 |
| Train4 | 546 | 43743 | 64 | 562 | >9999 | 62 | 64 | 572 | 585 | 530 |
| Train5 | 707 | 144339 | 242 | 2248 | >9999 | 243 | 243 | 2249 | 2251 | 981 |
| Train6 | 1164 | 29465 | 276 | 2083 | >9999 | 271 | 272 | 2102 | 2107 | 130 |
| Train7 | 1495 | 28639 | 349 | 2661 | >9999 | 343 | 344 | 2662 | 2673 | 358 |
| Train8 | 1873 | 50000 | 395 | 3137 | >9999 | 390 | 392 | 3239 | 3260 | 986 |
| Tram1 | 553 | 6437 | 49 | 420 | 24 | 49 | 49 | 420 | 422 | 23 |
| Tram2 | 553 | 29500 | 49 | 409 | 139 | 49 | 50 | 414 | 416 | 59 |
| Mean | 694 | 33937 | 138 | 1143 | 222 | 137 | 138 | 1153 | 1158 | 280 |

Table 1. Test instances and solutions. IP = TRACS II, SAA = Self Adjusting Algorithm.

Table 1 shows the problem sizes, the results of a commercial ILP solver called TRACS II (Fores et al, 2002) and the results of SAA. Table 2 lists the comparative results of the Evolutionary Squeaky Wheel Optimisation against the ILP and the self-adjusting approach, respectively. It also lists the results of the original Squeaky Wheel Optimization (Squeaky Wheel Optimisation). Compared with the integer programme solutions, our Evolutionary Squeaky Wheel Optimisation solutions are 0.78% better in terms of the number of shifts, and are only 0.11% more expensive in terms of the total cost. However, our results are much faster in general, particularly for larger instances. Compared with the best-performed self-adjusting approach, the Evolutionary Squeaky Wheel Optimisation has achieved better solutions for all data instances with generally similar execution times.

| Data | SWO Shift (best) | SWO Cost (best) | SWO CPU (sec) | ESWO Shift (best) | ESWO Cost (best) | ESWO CPU (sec) | IP ESWO Δshift | IP ESWO Δcost |
|---|---|---|---|---|---|---|---|---|
| Bus1 | 37 | 321 | >999 | 34 | 292 | 121 | 0.0 | 1.25 |
| Bus2 | 36 | 319 | >999 | 35 | 291 | 61 | 2.94 | 0.52 |
| Bus3 | 81 | 908 | >999 | 73 | 828 | 203 | -2.67 | -2.73 |
| Train1 | 67 | 554 | >999 | 62 | 507 | 141 | 0.00 | 0.39 |
| Train2 | 124 | 1097 | >999 | 116 | 994 | 176 | 0.00 | -0.92 |
| Train3 | 56 | 455 | >999 | 50 | 403 | 19 | 0.00 | -0.17 |
| Train4 | 67 | 632 | >999 | 61 | 569 | 536 | -4.69 | 1.25 |
| Train5 | 262 | 2488 | >999 | 242 | 2248 | 873 | 0.00 | 0.04 |
| Train6 | 314 | 2410 | >999 | 270 | 2082 | 135 | -2.17 | -0.04 |

| | | | | | | | |
|---|---|---|---|---|---|---|---|
| Train7 | 399 | 3091 | >999 | 342 | 2662 | 318 | -2.01 | 0.02 |
| Train8 | 447 | 3686 | >999 | 389 | 3200 | 928 | -1.52 | 1.99 |
| Tram1 | 53 | 444 | >999 | 49 | 420 | 27 | 0.00 | 0.00 |
| Tram2 | 54 | 437 | >999 | 49 | 411 | 74 | 0.00 | 0.56 |
| Mean | 154 | 1295 | >999 | 136 | 1147 | 278 | -0.78 | 0.11 |

Table 2. Comparative results. SWO = Squeaky Wheel Optimisation, ESWO = Evolutionary Squeaky Wheel Optimisation, IP = TRACS II.

## 4 Evolutionary Squeaky Wheel Optimisation for Nurse Scheduling

### 4.1 Problem description

Nurse scheduling has been widely studied in recent years, and a general overview of various approaches can be found in Cheng et al (2003), Burke et al (2004) and Rodrigues (2003). The nurse scheduling problem we consider in this paper is to create weekly schedules in a large UK hospital (Aickelin and Dowsland, 2000). The schedules must be seen 'fair' by the staff, which means unattractive patterns or weekend work has to be evenly spread. To achieve this all patterns carry a 'preference cost' (from zero = perfect to 100) and the objective is to minimize this. Furthermore, all schedules have to conform to a number of constraints regarding demand and skill levels. Full details can be found in (Aickelin and Dowsland, 2000). Typical problem dimensions are 30 nurses, three grade levels and over 400 theoretic weekly shift patterns per nurse. The integer programme model of the problem is shown below.

Decision variables:
$x_{ij} = 1$ if nurse i works shift pattern j, 0 otherwise.

Parameters:
m = Number of possible shift patterns;
n = Number of nurses;
p = Number of grades;
$a_{jk} = 1$ if shift pattern j covers day/night k, 0 otherwise;
$q_{ig} = 1$ if nurse i is of grade g or higher, 0 otherwise;
$c_{ij}$ = Preference cost of nurse i working shift pattern j;
$R_{kg}$ = Demand of nurses with grade g on day/night k;

Objective function:
$$\text{Minimize} \sum_{i=1}^{n} \sum_{j=1}^{m} c_{ij} x_{ij} \tag{15}$$

Subject to:
$$\sum_{j=1}^{m} x_{ij} = 1, \forall i \in \{1,...,n\} \tag{16}$$

$$\sum_{j=1}^{m}\sum_{i=1}^{n} q_{ig} a_{jk} x_{ij} \geq R_{ks}, \forall k \in \{1,...,14\}, g \in \{1,2,3\} \quad (17)$$

The goal of the objective function (15) is to minimize the total preference cost of all nurses. Constraint (16) ensures that every nurse works exactly one shift pattern from his/her feasible set, and constraint (17) ensures that the demand for nurses is fulfilled for every grade on every day and night. This problem can be regarded as a multiple-choice set-covering problem, in which the sets are given by the shift pattern vectors and the objective is to minimize the cost of the sets needed to provide sufficient cover for each shift at each grade. The constraints described in (16) enforce the choice of exactly one pattern (set) from the alternatives available for each nurse.

## 4.2 Application

This section presents the Evolutionary Squeaky Wheel Optimisation for nurse scheduling. Starting from a random initial solution, the steps described in section 4.2.1 to 4.2.5 are executed in sequence in a loop until a user specified parameter (e.g. CPU-time or solution quality) is reached or no improvement has been achieved for a certain number of iterations.

### 4.2.1 Analysis
This step evaluates the fitness of each nurse's assignment by taking the current whole schedules into account. The evaluating function used should be normalized. It should be able to evaluate the contribution of this assignment towards the cost and towards the feasibility aspect of the solution, formulated as:

$$F(s_i) = \frac{\max(c_{1j},...,c_{ij}) - c_{ij}}{\max(c_{1j},...,c_{ij}) - \min(c_{1j},...,c_{ij})} \bowtie \frac{\max(p_{1j},...,p_{ij}) - p_{ij}}{\max(p_{1j},...,p_{ij}) - \min(p_{1j},...,p_{ij})}, \forall i \in \{1,...,|S|\} \quad (18)$$

where S denotes the current schedule, $s_i$ denotes the shift pattern assigned to the i-th nurse, $c_{ij}$ is the preference cost of nurse i working shift pattern j, and $p_{ij}$ is the contribution towards the reduction in shortfall of qualified nurses for nurse i working shift pattern j.

In equation (19), $p_{ij}$ is calculated as the sum of grade one, two and three covered shifts that would become uncovered if the nurse does not work on this shift pattern. It can be formulated as:

$$p_{ij} = \sum_{g=1}^{3} q_{ig} (\sum_{k=1}^{14} a_{jk} d_{kg}) \quad (19)$$

where $q_{ig}$ and $a_{jk}$ use the same definitions as in the inequality represented in (17), $d_{kg}$ is 1 if there is a shortage of nurses during period k of grade g (i.e. the coverage value without considering shift pattern j is smaller than demand $R_{kg}$), 0 otherwise.

### 4.2.2 Selection
This step is to determine whether a shift pattern $s_i$, $i \in \{1,...,|S|\}$, should be retained or discarded. The decision is made by comparing its fitness value $F(s_i)$ to a random number generated for each iteration in the range [0, 1]. If $F(s_i)$ is larger, then $s_i$ will remain in its present allocation, otherwise $s_i$ will be removed from the current schedule. The days and nights that $s_i$ covers are then released and the coverage demands are updated. By using this kind of Selection, a shift pattern $S_i$ with larger fitness value $F(s_i)$ has a higher probability to survive in the current schedule.

### 4.2.3 Mutation

The Mutation step alters the shift patterns of the remaining nurses, i.e. it randomly discards them from the partial schedule at a small given rate $p_M$. The days and nights that a mutated $s_i$ covers are then released and coverage demands are updated.

### 4.2.4 Prioritization

The aim of the Prioritization step is to modify the previous sequence by generating a new priority sequence for the nurses that are waiting to be rescheduled (i.e. the ones that have been removed by the steps of Selection and Mutation), with poorly scheduled nurses being earlier in the sequence. Using the results of Analysis, this step first sorts the problem (or removed) shift patterns in ascending order of their fitness values. Since each shift pattern in the sequence corresponds to an associated nurse, we can then obtain from this list a sequence of problem nurses.

### 4.2.5 Construction

The constructing heuristic builds and/or repairs a schedule by assigning one of the shift patterns to each unscheduled nurse, in the order that the nurses appear in the priority sequence. A new schedule is obtained after each unscheduled nurse has been assigned a new shift pattern.

Based on the domain knowledge of nurse rostering, there are many rules that can be used to build schedules. Here we use three rules proposed in Aickelin et al (2007). They are alternatively used in each step of construction at the rate of $p_i$ (i=1, 2, 3) respectively, satisfying

$$\sum_{i=1}^{3} p_i = 1, p_i \geq 0 \tag{20}$$

The first rule is the "k-Cheapest" rule. Ignoring the feasibility of a solution, it randomly selects a shift pattern from a k-length list containing patterns with the k-cheapest costs $p_{ij}$, in an effort to reduce the schedule cost as much as possible.

The second, "Overall Cover" rule, is designed to consider only the feasibility of the schedule. It schedules one nurse at a time in such a way as to cover those days and nights with the largest amount of overall under cover, which is the sum of individual under cover of each shift.

The third "Combined" rule cycles through all possible shift patterns of a nurse, assigns each one a score and chooses the one with the highest score. In the case of more than one shift pattern with the best score, the first such shift pattern is chosen. The score $s_{ij}$ of shift pattern j for nurse i is calculated as the weighted sum of the nurse's $c_{ij}$ value for that particular shift pattern and its contribution to the cover of all three grades:

$$s_{ij} = w_p(100 - c_{ij}) + \sum_{g=1}^{3} w_g q_{ig} (\sum_{k=1}^{14} a_{jk} d_{kg}), \forall i \in \{1,...,m\}, j \in \{1,...,n\} \tag{21}$$

where $w_p$ is the weight of the nurse's $c_{ij}$ value for the shift pattern and $w_g$ is the weight of covering an uncovered shift of grade g. $a_{jk}$ and $q_{ig}$ are defined the same as in formula (17). $d_k$ is the number of nurses required if there are still nurses needed on day k of grade g, and 0 otherwise. (100−$c_{ij}$) is used in the score as higher $c_{ij}$ values are worse and the maximum for $c_{ij}$ is 100.

4.3 Experimental results

This section describes the computational experiments of the Evolutionary Squeaky Wheel Optimisation. For all experiments, the same 52 data instances are used. Each data instance consists of one week's requirements for all shift and grade combinations and a list of available nurses together with their preference costs and qualifications. A zip file containing all these 52 instances is available to download at http://www.cs.nott.ac.uk/~jpl/Nurse_Data/NurseData.zip.

In addition to the ILP approach (Aickelin and Dowsland, 2000), a variety of recently developed meta-heuristic techniques have also used the 52 instances as a benchmark testbed. The meta-heuristic approaches include a basic GA, an adaptive GA, a multi-population GA, a hill-climbing GA (all from Aickelin and White, 2004), an indirect GA (Aickelin and Dowsland, 2003), an estimated distribution algorithm (Aickelin and Li, 2006) and a learning classifier system (Li and Aickelin, 2004)), among which the indirect GA performs best and the hill-climbing GA second. All of these meta-heuristics use the following same objective function, retained here for a fair comparison:

$$\text{Minimize} \sum_{i=1}^{n}\sum_{j=1}^{m} c_{ij} x_{ij} + w_{demand} \sum_{k=1}^{14}\sum_{g=1}^{3} \max\left[ R_{kg} - \sum_{i=1}^{n}\sum_{j=1}^{m} q_{ig} a_{jk} x_{ij} ; 0 \right] \qquad (22)$$

Our Evolutionary Squeaky Wheel Optimisation approach was coded in Java 2 and implemented on the same Pentium 2.1 GHz PC used for the driver scheduling experiments. We start the search from an initial solution generated by assigning shift patterns to nurses at random, and set the stopping criterion as no improvement for 10,000 iterations or a known optimal solution has been found. The other parameters are set as follows: the mutation rate $p_M$ in Section 4.2.3 is 0.05, k of the "k-Cheapest" rule in Section 4.2.5 is 3, rule rate $p_c$ (i=1, 2, 3) in equation (20) is (0.02, 0.18, 0.80), weight distribution ($w_p$, $w_1$, $w_2$, $w_3$) in equation (21) is (1, 8, 2, 1), and penalty $w_{demand}$ in function (22) is 200. Note that, for consistency, these parameter values are the same as those used in previous papers solving the same 52 data instances.

Table 3 lists the optimal results found by a commercial ILP solver called XPRESS MP (Dowsland and Thompson, 2000). Apart from the summary results of the original Squeaky Wheel Optimisation, it also lists the comparative results of the Evolutionary Squeaky Wheel Optimisation against the best two meta-heuristic approaches (i.e. the indirect GA [IGA] and the hill-climbing GA [HGA]), based on 20 runs of each data instance using different random number seeds. These comparative results include the best, the mean and the p-values of t-test. The p-value indicates the significance of the hypothesis that, on average, the Evolutionary Squeaky Wheel Optimisation generates better solutions than the other approach for a given data instance at a 95% confidence level (p=0.05).

| Data | ILP | HGA Best | HGA Mean | IGA Best | IGA Mean | SWO Best | SWO Mean | ESWO Best | ESWO Mean | p-value HGA | p-value IGA |
|------|-----|----------|----------|----------|----------|----------|----------|-----------|-----------|-------------|-------------|
| 01 | 8 | 8 | 10 | 8 | 8 | 10 | 17 | 8 | 8 | 0.008 | 0.153 |
| 02 | 49 | 50 | 52 | 51 | 60 | 66 | 82 | 52 | 56 | 0.000 | 0.004 |
| 03 | 50 | 50 | 52 | 51 | 61 | 62 | 124 | 50 | 51 | 0.105 | 0.000 |
| 04 | 17 | 17 | 18 | 17 | 17 | 17 | 21 | 17 | 17 | 0.023 | 1.000 |
| 05 | 11 | 11 | 13 | 11 | 11 | 11 | 40 | 11 | 11 | 0.110 | 0.324 |
| 06 | 2 | 2 | 5 | 2 | 2 | 2 | 12 | 2 | 2 | 0.026 | 0.075 |
| 07 | 11 | 13 | 139 | 12 | 22 | 12 | 110 | 11 | 12 | 0.000 | 0.000 |
| 08 | 14 | 14 | 16 | 15 | 19 | 27 | 39 | 14 | 14 | 0.003 | 0.000 |
| 09 | 3 | 3 | 40 | 4 | 6 | 4 | 23 | 3 | 4 | 0.013 | 0.031 |
| 10 | 2 | 2 | 9 | 3 | 4 | 12 | 21 | 2 | 3 | 0.000 | 0.000 |
| 11 | 2 | 2 | 5 | 2 | 3 | 13 | 67 | 2 | 2 | 0.005 | 0.027 |
| 12 | 2 | 2 | 26 | 2 | 2 | 6 | 117 | 2 | 2 | 0.010 | 1.000 |
| 13 | 2 | 2 | 3 | 2 | 2 | 6 | 81 | 2 | 2 | 0.028 | 0.768 |
| 14 | 3 | 3 | 5 | 3 | 8 | 4 | 99 | 3 | 3 | 0.020 | 0.000 |
| 15 | 3 | 3 | 6 | 3 | 5 | 4 | 9 | 3 | 4 | 0.066 | 0.000 |
| 16 | 37 | 38 | 98 | 39 | 44 | 46 | 106 | 37 | 38 | 0.000 | 0.000 |
| 17 | 9 | 9 | 22 | 10 | 15 | 13 | 125 | 9 | 9 | 0.000 | 0.001 |
| 18 | 18 | 19 | 68 | 18 | 20 | 34 | 91 | 18 | 19 | 0.001 | 0.000 |
| 19 | 1 | 1 | 15 | 1 | 2 | 5 | 75 | 1 | 1 | 0.021 | 0.033 |
| 20 | 7 | 8 | 41 | 7 | 15 | 11 | 55 | 7 | 8 | 0.009 | 0.000 |
| 21 | 0 | 0 | 10 | 0 | 0 | 2 | 28 | 0 | 0 | 0.011 | 1.000 |
| 22 | 25 | 25 | 32 | 25 | 26 | 37 | 143 | 25 | 26 | 0.022 | 0.846 |
| 23 | 0 | 0 | 48 | 0 | 1 | 102 | 198 | 0 | 0 | 0.025 | 0.000 |
| 24 | 1 | 1 | 84 | 1 | 1 | 10 | 33 | 1 | 1 | 0.003 | 0.154 |
| 25 | 0 | 0 | 3 | 0 | 0 | 4 | 94 | 0 | 0 | 0.139 | 0.324 |
| 26 | 48 | 48 | 204 | 48 | 132 | 250 | 305 | 48 | 49 | 0.000 | 0.001 |
| 27 | 2 | 2 | 30 | 4 | 14 | 204 | 240 | 2 | 3 | 0.003 | 0.000 |
| 28 | 63 | 63 | 68 | 64 | 150 | 81 | 104 | 63 | 64 | 0.081 | 0.000 |
| 29 | 15 | 17 | 216 | 15 | 27 | 15 | 117 | 15 | 16 | 0.000 | 0.365 |
| 30 | 35 | 35 | 42 | 38 | 41 | 65 | 143 | 35 | 37 | 0.398 | 0.000 |
| 31 | 62 | 161 | 237 | 65 | 77 | 152 | 171 | 65 | 70 | 0.000 | 0.000 |
| 32 | 40 | 41 | 90 | 42 | 46 | 52 | 177 | 40 | 41 | 0.000 | 0.000 |
| 33 | 10 | 12 | 45 | 12 | 15 | 104 | 106 | 10 | 11 | 0.046 | 0.000 |
| 34 | 38 | 40 | 107 | 39 | 44 | 77 | 165 | 38 | 40 | 0.000 | 0.000 |
| 35 | 35 | 35 | 36 | 36 | 44 | 77 | 106 | 35 | 36 | 0.002 | 0.000 |
| 36 | 32 | 33 | 77 | 32 | 37 | 51 | 83 | 32 | 34 | 0.006 | 0.020 |
| 37 | 5 | 5 | 10 | 5 | 9 | 23 | 73 | 5 | 6 | 0.083 | 0.000 |
| 38 | 13 | 16 | 96 | 15 | 19 | 29 | 107 | 13 | 15 | 0.000 | 0.000 |
| 39 | 5 | 5 | 8 | 5 | 8 | 13 | 90 | 5 | 6 | 0.286 | 0.028 |
| 40 | 7 | 8 | 16 | 7 | 10 | 20 | 68 | 7 | 8 | 0.003 | 0.009 |
| 41 | 54 | 54 | 109 | 55 | 68 | 177 | 203 | 54 | 54 | 0.002 | 0.000 |
| 42 | 38 | 38 | 48 | 39 | 44 | 39 | 224 | 38 | 39 | 0.001 | 0.005 |
| 43 | 22 | 24 | 99 | 23 | 24 | 42 | 132 | 22 | 22 | 0.000 | 0.000 |
| 44 | 19 | 48 | 117 | 25 | 31 | 28 | 110 | 19 | 21 | 0.000 | 0.000 |
| 45 | 3 | 3 | 5 | 3 | 5 | 23 | 63 | 3 | 4 | 0.040 | 0.193 |
| 46 | 3 | 6 | 59 | 6 | 8 | 217 | 323 | 3 | 5 | 0.002 | 0.000 |
| 47 | 3 | 3 | 9 | 3 | 4 | 39 | 221 | 3 | 3 | 0.000 | 0.006 |
| 48 | 4 | 4 | 12 | 4 | 9 | 33 | 124 | 4 | 5 | 0.004 | 0.000 |
| 49 | 27 | 29 | 104 | 30 | 233 | 122 | 136 | 28 | 31 | 0.000 | 0.000 |

| 50 | 107 | 110 | 223 | 110 | 241 | 109 | 189 | 108 | 110 | 0.000 | 0.000 |
| 51 | 74 | 75 | 246 | 74 | 93 | 175 | 207 | 74 | 107 | 0.000 | 0.373 |
| 52 | 58 | 75 | 169 | 58 | 67 | 113 | 179 | 59 | 65 | 0.000 | 0.293 |
| mean | 21 | 24 | 63 | 22 | 36 | 55 | 115 | 21 | 23 | n/a | n/a |

Table 3. Comparative results (italics indicate that the results using the Evolutionary Squeaky Wheel Optimisation are significantly better)

From Table 3, we can also see how the original Squeaky Wheel Optimisation is far from sufficient to solve the problem, how the Evolutionary Squeaky Wheel Optimisation performs as well as the ILP approach (with 47 of 52 instances being optimal and the rest 5 near-optimal) and how the Evolutionary Squeaky Wheel Optimisation obtains much better results than the best indirect GA and the second best hill-climbing GA. Compared with the hill-climbing GA, the Evolutionary Squeaky Wheel Optimisation performs better in 51 out of 52 instances, in which the difference is statistically significant for 43 instances. Compared with the indirect GA, for 38 of 52 instances the performance of the Evolutionary Squeaky Wheel Optimisation is statistically better. Out of the remaining 14 instances, in seven of them (i.e. instances 01, 04, 05, 06, 12, 24 and 25) the performance of the Evolutionary Squeaky Wheel Optimisation is 100% optimal or very nearly optimal (with only 1 of 20 runs being non-optimal), thus there is no difference between the Evolutionary Squeaky Wheel Optimisation and the indirect GA. For the five instances of 21, 24, 29, 45 and 52, the performance of the Evolutionary Squeaky Wheel Optimisation is better but the difference is not statistically significant. Our Evolutionary Squeaky Wheel Optimisation is slightly outperformed by the indirect GA in only 2 out of 52 instances.

In terms of the CPU time, the Evolutionary Squeaky Wheel Optimisation takes between 1-200 seconds per run and per data instance depending on the difficulty of individual instances, with an average of 39 seconds. Table 4 lists the average runtimes of various approaches over these 52 instances. For more than half of the instances, an optimal solution can be found within 30 seconds. This is much shorter than the 24 hours runtime of the ILP approach, and, on average, a little longer than that of the hill-climbing GA and the indirect GA. Although the times below have been adjusted to allow for the different hardware platforms they were derived from, a direct comparison is not possible between the heuristics as they were programmed in different languages and styles. For instance, the IGA was written in Pascal, the HGA in C and SWO / ESWO in Java. Moreover, both HGA and IGA had no user interface and were optimised for off-line experiments, whereas SWO/ESWO has options for graphical input and output. Thus overall, we would argue their performance is similar.

|  | IP | HGA | IGA | SWO | ESWO |
|---|---|---|---|---|---|
| Time (sec) | >24hours | 15 | 12 | 39 | 39 |

Table 4. Comparison of the average runtime of various approaches

Our superior results reported in Table 3 are obtained by alternately using three rules at different rates for solution construction. Naturally, one might be interested to know the relative gain achieved by each of these rules. Or in other words, what happens if don't use any rule at all (i.e., assign shift patterns to nurses at random), and what happens in the three extreme cases produced by giving weight only to rule 1, 2 and 3 in equation (20), respectively (i.e., by choosing $p_i = 1$ for $i = 1, 2$ and $3$). Table 5 gives such results, from which we find that neither "Random" assignment nor the "k-Cheapest" rule can produce any feasible solutions (i.e., the ones with the cost values smaller than 400), the "Overall Cover" rule performs better, and the "Combined" rule performs the best.

| Data | Random | k-cheapest | Overall cover | Combined | Data | Random | k-cheapest | Overall cover | Combined |
|---|---|---|---|---|---|---|---|---|---|
| 01 | 1041 | 703 | 42 | 8 | 27 | 1512 | 1786 | 129 | 47 |
| 02 | 730 | 1012 | 129 | 58 | 28 | 1295 | 1812 | 229 | 85 |
| 03 | 1090 | 1953 | 117 | 54 | 29 | 1347 | 1058 | 172 | 136 |
| 04 | 576 | 1398 | 51 | 17 | 30 | 1290 | 519 | 213 | 204 |
| 05 | 943 | 890 | 78 | 11 | 31 | 1168 | 1039 | 220 | 368 |
| 06 | 1155 | 1563 | 169 | 2 | 32 | 1164 | 1324 | 143 | 82 |
| 07 | 714 | 1444 | 164 | 68 | 33 | 1125 | 1720 | 127 | 36 |
| 08 | 1208 | 1433 | 103 | 15 | 34 | 916 | 926 | 142 | 117 |
| 09 | 1051 | 997 | 99 | 11 | 35 | 1222 | 793 | 147 | 48 |
| 10 | 1367 | 1155 | 101 | 3 | 36 | 1009 | 384 | 157 | 39 |
| 11 | 1176 | 666 | 49 | 3 | 37 | 1309 | 1082 | 64 | 11 |
| 12 | 1135 | 1135 | 34 | 23 | 38 | 1094 | 954 | 70 | 22 |
| 13 | 982 | 1530 | 66 | 3 | 39 | 1012 | 972 | 50 | 8 |
| 14 | 1243 | 1100 | 218 | 3 | 40 | 1146 | 1338 | 148 | 8 |
| 15 | 927 | 684 | 117 | 4 | 41 | 1143 | 1817 | 253 | 353 |
| 16 | 1075 | 1772 | 185 | 237 | 42 | 1369 | 855 | 239 | 43 |
| 17 | 1214 | 1092 | 180 | 11 | 43 | 1416 | 1155 | 186 | 84 |
| 18 | 1310 | 1229 | 169 | 18 | 44 | 1140 | 1625 | 180 | 31 |
| 19 | 1066 | 2331 | 132 | 22 | 45 | 1091 | 428 | 60 | 6 |
| 20 | 1137 | 1567 | 144 | 14 | 46 | 1552 | 1298 | 112 | 8 |
| 21 | 1088 | 457 | 30 | 1 | 47 | 1610 | 2100 | 89 | 5 |
| 22 | 1198 | 1569 | 83 | 31 | 48 | 1152 | 1956 | 103 | 7 |
| 23 | 1485 | 2637 | 87 | 1 | 49 | 1391 | 1155 | 145 | 189 |
| 24 | 1489 | 2549 | 76 | 1 | 50 | 1312 | 1328 | 305 | 276 |
| 25 | 1063 | 2906 | 25 | 0 | 51 | 1573 | 1493 | 252 | 430 |
| 26 | 1592 | 3284 | 185 | 229 | 52 | 1057 | 1822 | 133 | 123 |

Table 5. Average results of 20 runs by random assignment and by single rules

## 5 Discussion and Conclusions

This paper presents a new technique to solve personnel scheduling problems by using the original idea of Squeaky Wheel Optimisation but by incorporating two new operators. With these two additional steps, the drawbacks of the original Squeaky Wheel Optimisation in terms of its search ability and execution speed are successfully dealt with. Taken as a whole, the Evolutionary Squeaky Wheel Optimisation implements evolution on a single solution and undertakes search via iterative disruption, improvement and construction.

So how far have we succeeded in our goal of developing a robust heuristic? The experiments have demonstrated that the Evolutionary Squeaky Wheel Optimisation performs very efficiently and competitively on a range of problem instances in the two well-studied domains of transportation driver scheduling and hospital nurse scheduling. In general, it outperforms the previous best-performing approaches reported in the literature. Although broad similarities exist, the fact that these domains are actually very different supports our claim that the Evolutionary Squeaky Wheel Optimisation is an efficient technique that can be employed across different versions of the personnel scheduling problem.

The architecture of the Evolutionary Squeaky Wheel Optimisation is innovative and there is still some room for further improvement. For example, although results for our current test instances are very good, as we only use a very limited number of rules, we believe that by adding some more rules into the search, solution quality could be improved further. This would be particularly interesting if we have more difficult instances to solve. In the future, we are also looking at more advanced methods of Analysis, Selection and Mutation. In addition, we are looking at other domains that would enable us to further demonstrate the generality of the technique.

## Acknowledgements

The research described in this paper was funded by the Engineering and Physical Sciences Research Council (EPSRC), under grant GR/S70197/1.